\documentclass[sigconf]{acmart}

 \setcopyright{none}
\renewcommand\footnotetextcopyrightpermission[1]{} 
\usepackage[lined,ruled,boxed,commentsnumbered,linesnumbered]{algorithm2e}
\usepackage{bbm}
\include{figure}
\include{caption}
\usepackage{dblfloatfix} 
\newtheorem{Def}{Definition}
\newtheorem{Prop}{Proposition}
\newtheorem{Thm}{Theorem}
\begin{document}

\title{Unsupervised Early Exit in DNNs with Multiple Exits }


\author{Hari Narayan N U}
\affiliation{%
  \institution{MLiONS Lab, IEOR}
  \city{IIT Bombay, Mumbai}
  \country{India}}
\email{203170005@iitb.ac.in}

\author{Manjesh K. Hanawal}
\affiliation{%
  \institution{MLiONS Lab, IEOR}
  \city{IIT Bombay, Mumbai}
  \country{India}}
  \email{mhanawal@iitb.ac.in}

  \author{Avinash Bhardwaj}
\affiliation{%
  \institution{Mechanical Engineering}
  \city{IIT Bombay, Mumbai}
  \country{India}}
  \email{abhardwaj@iitb.ac.in}

\begin{abstract}
Deep Neural Networks (DNNs) are generally designed as sequentially cascaded differentiable blocks/layers with a prediction module  
connected only to its last layer.  DNNs can be attached with prediction modules at multiple points along the backbone where inference can stop at an intermediary stage without passing through all the modules. The last exit point may offer a better prediction error but also involves more computational resources and latency. An exit point that is `optimal' in terms of both prediction error and cost is desirable. The optimal exit point may depend on the latent distribution of the tasks and may change from one task type to another. During neural inference, the ground truth of instances may not be available and hence the error rates at each exit point cannot be estimated. Hence one is faced with the problem of selecting the optimal exit in an unsupervised setting.  Prior works tackled this problem in an offline supervised setting assuming that enough labeled data is available to estimate the error rate at each exit point and tune the parameters for better accuracy. However, pre-trained DNNs are often deployed in new domains for which a large amount of ground truth may not be available.  We thus model the problem of exit selection as an unsupervised online learning problem and leverage the bandit theory to identify the optimal exit point. Specifically, we focus on the Elastic BERT, a pre-trained multi-exit DNN to demonstrate that it `nearly' satisfies the Strong Dominance (SD) property  making it possible to learn the optimal exit in an online setup without knowing the ground truth labels. We develop upper confidence bound (UCB) based algorithm named UEE-UCB that provably achieves sub-linear regret under the SD property. Thus our method provides a means to adaptively learn domain-specific optimal exit points in multi-exit DNNs. We empirically validate our algorithm on IMDb and Yelp datasets.

\end{abstract}

\maketitle
\pagestyle{empty}

\section{Introduction}
The success of modern neural networks is partly attributed to their deep nature (a large number of sequentially cascaded layers). This has led to improved accuracies across benchmark datasets at the cost of increased latency and computation costs. However, latency and computational cost are major concerns when it comes to real-world model deployment. Consider the case where we need to perform model inference in mobile or edge devices. In such situations, the lack of appropriate computing facilities makes it hard to run model inference entirely on the device. An alternative approach is to implement the initial layers of DNN in the local devices and the remaining layers in a remote high-resource device. The inference process can connect to the remote device only if the quality of the inference from the layers in the local devices is not good enough. Otherwise, inference can be done locally without incurring additional costs.

Muti-exit DNNs are well suited for mobile-cloud inference, where the early stages of the DNNs are processed in the mobile, and the rest are offloaded to the cloud. In multi-exit DNNs, multiple exits are trained along the backbone neural architecture. When an input sample is fed to such networks, an output prediction is available at each of these exits. The initial exits have low computational costs but high error probabilities, while deeper exit points will have lower error probabilities but higher costs. Once such a multiple exit model is trained, during the inference stage, we are faced with an unsupervised problem of selecting an optimal exit point for each input sample based on the cost vs error trade-off. We refer to the weighted sum of error rates and cost of exit points as its loss and our interest is to find an exit point with minimal loss.

One issue with using any pre-trained models, like multi-exit DNNs, is that they are often used as black boxes with fixed thresholds. For example, samples are passed sequentially through the exit points. At each exit point entropy values are compared against a fixed threshold to decide whether to exit or continue processing the sample. Smaller values of the threshold at each exit drive samples to exit early while larger values make them use later exit points. The best threshold can depend on the distributions of the samples which could be unknown a priori.  As multi-exit DNNs are often used as pre-trained models to apply to scenarios where the distribution of the samples could be different from that of the training phase, the threshold needs to be adapted to the new domain to achieve the best performance.
This necessitates online learning for exit selection based on the observations from the new domain samples. 

The task of training a multi-exit model is subject to various choices such as the optimal placing of exit branches, the exit branch architecture, training strategy of exits \cite{bapna2020controlling, ACL2020_Deebert, kaya2019shallow} etc. There is also the question of how to evaluate the performance of such models  \cite{Scardapane_2020}. In this paper, we model the problem of exit selection as an online learning setup that provides a framework for evaluating the performance of the inference stage using the notion of cumulative regret. In the inference stage of the deployed DNNs, ground truth labels may not be available and hence one cannot verify if the inferences of the exits points are correct and hence cannot estimate the errors rates of the exit points leading to an unsupervised online learning setup or more generally known as partial monitoring setups \cite{partialmonitoring}. 

The performance of any online algorithm depends on the quality of the feedback received which helps estimate the quantities of interest. In the unsupervised online learning setup, the feedback at each round is the inference made by the exit points. Due to the non-availability of the ground truths this feedback cannot  be used to estimate the error rate of the exit points and one cannot learn the optimal exit point. However, what could come to one's advantage is the specific problem structure that could help identify the optimal arm without knowing the true error rate or loss associated with the exit points. Unsupervised Sensor Section (USS) setup introduced in \cite{hanawal2017unsupervised, verma2019Unsupervised} provides conditions on the problem structure under which optimal action can be identified. Specifically, one of the conditions states when the problem instance satisfies the Strong Dominance (SD) property, one can identify the optimal action. The inverse of this property is noted in \cite{kaya2019shallow}  as overthinking in neural networks.

SD property states that if an exit point in a multi-exit DNN makes a correct inference, all the later exit points also make a correct inference with probability one. By  design, multi-exit DNNs are expected to 'largely' satisfy this property as additional layers extract more refined features which can only improve their inference performance. In case the SD property is not satisfied by all samples, there will be a small penalty which is insignificant as we show later in the experiment sections.

Attention-based language models are natural test beds for multi-exit network research. We use Elastic-BERT \cite{liu2021elasticbert} which is a state-of-the-art multi-exit DNN for natural language inference to demonstrate the efficacy of our method. Elastic-BERT uses the BERT \cite{BERTPretrain} as a backbone which is trained with multiple exit points along the backbone on a large corpus of natural language datasets. We use Elastic-BERT for the task of sentiment classification on IMDb \cite{IMDb} and Yelp \cite{Yelp} datasets.  


Our contributions can be summarized as follows: 
\begin{itemize}
    \item We model exit selection in neural networks as an unsupervised online learning problem.
    \item We develop an upper confidence-based algorithm named UEE-UCB to identify the optimal exit point and show that it achieves sublinear regret under the Strong Dominance property.
    \item We empirically validate that the Strong Dominance property `nearly' holds across the cascade of exit points of ElasticBERT \cite{liu2021elasticbert} model for IMDb and Yelp datasets.
    \item We experimentally validate that UEE-UCB identifies the optimal exit on IMDb and Yelp datasets.
\end{itemize}

The rest of the paper is organized as follows: In section \ref{sec:RelatedWork}, we discuss the related works and discuss how our work is different from the existing literature. In Section \ref{sec:ProblemSetup}, we discuss the problem setup, define objectives and state the assumptions. We develop an algorithm named UEE-UCB for learning optimal exit in multi-exit DNNs in Section \ref{sec:Algorithm}. In Section \ref{sec:TrainingDNN}, we discuss how we build a multi-exit DNN and experimentally validate the performance of UEE-UCB on this DNN. We give conclusions and future work directions in Section \ref{sec:Conclusion}. The code is available at \url{https://github.com/MLiONS/MutiExitDNNs}.

\section{Related Work}
\label{sec:RelatedWork}

In this section, we discuss the literature on multi-exit DNNs and the use of multi-armed bandits in multi-exit DNNs.

\subsection{Multi-Exit in Deep Neural Networks}
\label{sec:relatedEarly}

BranchyNet~\cite{teerapittayanon2016branchynet} is a multi-exit DNN that uses the classification entropy at each exit to decide whether a sample can be classified earlier, i.e., on a side branch. BranchyNet verifies whether the entropy value of the current prediction is greater than a predefined and fixed threshold. If so, the inference ends, and the sample is classified on the side branch. Two similar architectures, SPINN~\cite{laskaridis2020spinn} and SEE~\cite{wang2019see} make this decision based on the estimated classification confidence provided by a side branch. The confidence is given by the probability of the most likely class. 

Besides BranchyNet and SPINN, other works also employ multi-exit DNNs to reduce inference time. FlexDNN~\cite{fang2020flexdnn} and Edgent~\cite{li2019edge} use multi-exit DNNs to select the most appropriate DNN depth. 
Some works focus on deploying multi-exit DNNs in hardware. Dynexit~\cite{wang2019dynexit} trains and deploys a multi-exit DNN on Field Programmable Gate Array (FPGA) hardware. Meanwhile, Paul~\textit{et al.}~\cite{kim2020low} show that implementing a multi-exit DNN on the FPGA board can reduce inference time and energy consumption.

Pacheco~\textit{et al.}~\cite{pacheco2021calibration} combine multi-exit DNN and DNN partitioning to offload mobile devices via multi-exit DNNs. This offloading scenario is also considered in~\cite{pacheco2021distorted}, which proposes a robust multi-exit DNN against image distortion. In a similar vein, EPNet~\cite{dai2020epnet} learns when to multi-exit  accounting for the tradeoff between overhead and accuracy, but the learning occurs in an offline fashion. 

Muti-exit DNNs are being adapted in various other domains like  ranking systems \cite{WSDM2010_EarlyExitRanking}, Image classification \cite{ICLR2018_ResourceEfficient} and natural language processing \cite{bapna2020controlling,elbayad19arxiv, DBLP:conf/aclnmt/DabreRF20, liu2021elasticbert, ACL2020_Deebert}. DeeBERT \cite{ACL2020_Deebert} and ElasticBERT\cite{liu2021elasticbert} are based on the transformer-based BERT model. The DeeBERT is obtained by training the inference modules attached before the last module to the BERT backbone, whereas ElasticBERT is obtained by training both the BERT backbone with all the attached exit points.   

All multi-exit DNNs discussed above work with a fixed threshold on the exit points, which is not adapted to the task domain. Our work overcomes this limitation by learning domain-specific optimal exit using the online learning framework. 

\subsection{Multi-Armed Bandits in Multi-Exit DNNs} 

Most previous works decide the appropriate early exit based on entropy or confidence values and compare it to a fixed threshold. LEE~\cite{ju2021learning} and DEE~\cite{ju2021dynamic} are two notable exceptions that learn the optimal exit in a multi-exit DNN using multi-armed bandits. LEE and DEE aim to provide an efficient DNN inference task for mobile devices, e.g.,  during service outages and network disconnections. To this end, they consider a mobile-only scenario, in which the entire multi-exit DNN model is processed at a mobile or edge device. Although the motivation of our work is similar to~\cite{ju2021learning,ju2021dynamic}, our paper differs from it in at least three key aspects. Our problem formulation considers a weighted combination of accuracy and the cost of each exit point to decide where to exit. The cost in our setup is generic which could represent latency or computational cost. Both LEE and DEE assume that utility is revealed when an exit point is selected, which could depend on the ground truth labels. LEE uses the classical UCB1 \cite{ML02_UCB1_Auer} algorithm to learn the optimal exit point, and DEE uses the contextual bandit framework to adaptively find an optimal exit for each sample.  Our work is different from LEE and DEE as we do not observe any utility in our setup, and hence the feedback model is different. In our setup, only the prediction from the selected exit points is observed and no information is available about the ground truths.

Our work builds on the Unsupervised Sensor Selection (USS) framework developed in \cite{hanawal2017unsupervised} for learning optimal exit points in multi-exit DNNs. We exploit the structural property of the multi-exit DNNs to develop an algorithm that  has sublinear regret. Also, previous works focused on mobile devices, our approach is generic. In this work, we focus specifically on multi-exit DNNs based on BERT model to demonstrate the effectiveness of our method.

\section{Problem Setup}
\label{sec:ProblemSetup}
We are given a pre-trained multi-exit DNN (for classification task) along with the cost associated with each exit. Let $K$ denote the number of exits in the multi-exit DNN. Each exit $k\in [K]$, where $[K]=\{1,2,...,K\}$, is associated with an error rate defined as $\gamma_k=\Pr\{Y\neq \hat{Y}_k\}$, where $\hat{Y}_k$ is the inference made by exit $k$ when the ground truth label is $Y$. This error rate depends on the probability space from which the samples are drawn and their associated latent label generation process. In our setup, the labels associated with the samples are not revealed; hence, $\gamma_k$ for any $k\in [K]$ cannot be estimated. The usage of each exit point has a certain cost associated with it. This can be the actual computational cost or the latency in completing the inference. We denote the cost associated with the $k^{th}$ exit point as $c_k$. In multi-exit DNNs, each sample goes through the exit points sequentially and the cost is accumulated.  $c_k$ denotes the accumulated cost at the $k^{th}$ exit.  Let $l_k, k\in [K]$ denote the mean loss incurred when a sample exits from the $k^{th}$ exit point. It is defined as
\[l_k:=\gamma_k + \lambda c_k= \mathbb{E}\left [\mathbbm{1}{\{Y\neq \hat{Y}_k\}} \right ]+ \lambda c_k,\]
where $\hat{Y}_k$ denote inference made by the $k^{th}$ exit on a sample with ground label $Y$, and $\lambda$ denotes the proportionality constant that decides the trade-off between accuracy and cost. Without loss of generality, we set $\lambda = 1$ as its value can be absorbed with the cost values and rescaled. To keep the terminology consistent with the bandit literature, we refer to exit $k$ as arm $k$ and $l_k$ as the mean loss associated with it. In the following, we use the arm and exit point interchangeably. 

We assume that $\gamma_1\geq \gamma_2\geq \ldots \geq \gamma_K$, i.e., the error rates are decreasing as we move towards deeper exits. This assumption is natural in multi-exit DNNs as deeper exists will have access to more refined features which is likely to improve the inference performance on an average (not necessarily in a sample-wise sense). Indeed, 
this nearly holds on the IMDb and Yelp datasets as we show later (see Fig.2). Naturally, the costs on the exit points  are increasing, i.e., $c_1\leq c_2\ldots \leq c_K$. Note that the ordering of mean loss values $l_k$'s can be arbitrary and are unknown as $\gamma_k$'s are unknown. Our goal is to learn the arm with the lowest total cost, i.e., $k^*=\arg\min l_k$.  

We develop an online learning algorithm to identify the optimal exit. Let $\pi_t$ denote a policy that selects an exit point for a sample received in round $t$ based on past observations.  Let $I_t$ denote the exit selected by the policy. The expected cumulative regret of the policy $\pi=\{\pi_t: t \in [n]\}$ over $n$ round is defined as 
\begin{equation}
\label{eqn:Regret}
R(\pi, n)=\sum_{t=1}^n \mathbb{E}[l_{I_t}-l_{k^*}],
\end{equation}
where expectation with the respect to the randomness in the arm selection induced by the observed samples. We aim to develop a learning algorithms that gives sub-linear regret i.e., $R(\pi,n)/n \rightarrow 0$.

Notice that $l_k$'s cannot be estimated in our setup due to the non-availability of the labels. However, we have still set the ambitious goal of minimizing regret as in classical multi-armed bandits where the mean losses can be estimated. Clearly, any online algorithm will be unable to find an optimal arm if it cannot estimate the values of $l_k$'s. Hence we make some assumptions on the problem structure that enables us to learn an optimal arm. The Strong Dominance (SD) property introduced in \cite{verma2019Unsupervised, hanawal2017unsupervised} is one such property under which learning an optimal arm is possible in the unsupervised setting.

\begin{Def}[Strong Dominance]
We say that multi-exit DNN satisfies SD property if on any sample with ground truth label $Y$, the following holds:
\[ \hat{Y}_k=Y \;\;\mbox{for some}\;\; k\in [K] \implies \hat{Y}_j=Y \;\; \forall j>k. \]
\end{Def}

SD property requires that if an exit makes the correct inference, all the subsequent exits should also make the correct inference. SD property is expected to hold nearly for multi-exit DNNs as deeper exit points make inferences based on a richer set of features. However, we note that the SD property is a strong assumption and all the samples may not satisfy this property with probability one.  We only need SD property to develop a theoretically sound algorithm. As we will see later, our algorithm is built assuming the SD property, however, it works well even when the SD property is violated on a fraction of the samples. 

We note that whether or not the SD property holds depends on both multi-exit DNNs and the underlying distributions of the samples on which inference is performed. If the underlying sample generation process changes, the SD property may not continue to hold.  Also, the SD property is related to the over-thinking \cite{kaya2019shallow} phenomenon in neural networks.

Under the SD property, we have the following result that connects the difference in error rates of two exit points with their disagreement probabilities. 

\begin{Prop}
	Let $i,j\in [K]$ and $i<j$. Assume that the given multi-exit DNNs satisfy the SD property over the sample distributions. Then, the error rates induced by the sample distribution satisfy the following relation
	\[\gamma_i - \gamma_j =\Pr\{\hat{Y}_i \neq \hat{Y}_j\}.\]
\end{Prop}
The proof follows by applying SD property in \cite{hanawal2017unsupervised}[Prop. 3]. We next exploit this result to develop Upper Confidence Bound (UCB) based algorithm to learn the optimal exit points in multi-exit DNNs.

\section{Algorithm}
\label{sec:Algorithm}
When a sample is input to the multi-exit DNNs, we get the inference from all the exits it passes through. Let $(\hat{Y}_1,\hat{Y}_2\ldots, \hat{Y}_k)$
denote the predictions when the inference is terminated at the $k^{th}$ exit point. As discussed earlier, this information alone cannot be used to estimate the error rate. However, we can compare the predictions from the exit points and can use the outcome of comparisons to estimate disagreement probabilities. As we will argue next, under the SD property learning the disagreement probabilities is enough to learn the optimal arm. 

Define $r_k:=l_1 - l_k=\gamma_1 + c_1 - \gamma_k -c_k$. Then it is clear that $\arg\max r_k = \arg \min l_k$ and the relation $r_k= \Pr\{\hat{Y}_1\neq \hat{Y}_k\}+ c_1-c_k$ holds under the SD property. Though $l_k$ cannot be estimated for any $k$, one can estimate $\Pr\{\hat{Y}_1\neq \hat{Y}_k\}$ ( and $r_k$), by only looking at the disagreement between the prediction of $k^{th}$ exit point with that of the first exit point. Using the estimates of $r_k$'s, we can find the index of the arm that maximizes $r_k$'s which is also the index of the desired optimal arm.  Also, note the output of the first exit is always observed which makes it feasible to compare predictions of other exists used with that of the first exit in each round. We use this observation to develop an Upper Confidence Bound (UCB) based algorithm which we refer to as Unsupervised Early Exit (UEE-UCB).

\begin{algorithm}[t]
\caption{UEE-UCB} \label{alg:UEE-UCB}
\textbf{Input:} $\alpha>1$, $K$ (number of exits), costs $c_k \; \forall k \in [K]$ \\
\textbf{Initialize} $N_k\leftarrow 0, X_k \leftarrow 0$, $P_{1k} \leftarrow 0 \quad \forall k \in [K]$ \\
Use the last exit for the first sample \\ 
Observe $(\hat{Y}^1_1, \hat{Y}^1_2, \ldots, \hat{Y}^1_K)$\\
$N_k\leftarrow N_k+1 \quad \forall k \in [K]$ \\
$X_k\leftarrow X_k+ \mathbbm{1}{\{\hat{Y}^1_1 \neq \hat{Y}^1_k\}} \quad \forall k \in [K]$\\
$P_{1k} \leftarrow X_k/N_k \quad \forall k \in [K]$ \\
\For{\texttt{$t=2,3,\dots,$}}{
  Receive a sample \\
$I_t \leftarrow \textrm{argmax}_{k \in [K]} \left( P_{1k}+ c_1 -c_k + \sqrt \frac{\alpha \ln(t)}{ N_k}  \right)$\;
Use exit point $I_t$ \\
Observe $(\hat{Y}^t_1, \hat{Y}^t_2, \ldots, \hat{Y}^t_{I_t})$\\
$N_k\leftarrow N_k+1 \quad \forall k \in [I_t]$ \\
$X_k\leftarrow X_k+ \mathbbm{1}{\{\hat{Y}^t_1 \neq \hat{Y}^t_k\}} \quad \forall k \in [I_t]$\\
$P_{1k} \leftarrow X_k/N_k \quad \forall k \in [I_t]$ \\
}
\end{algorithm}
\begin{figure*}[btp]
	\label{fig:TrainDNN}
	\vspace{-1.2cm}
	\centering
	\includegraphics[scale=.9]{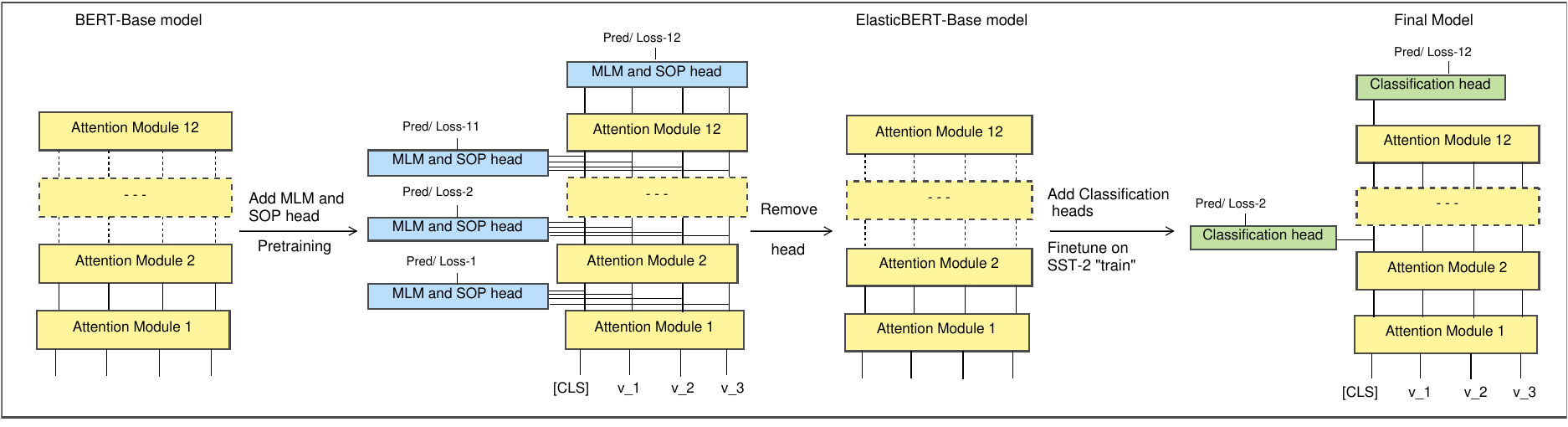}
	\caption{Procedure adopted in training Multi-exit neural network}
\end{figure*}

UEE-UCB takes $\alpha>1$ and $K$ as the input parameters. The parameter $\alpha$ decides the exploration rate. In the first round, the sample is taken till the last exit and predictions from all the exit points are observed, i.e., $(\hat{Y}_1,\hat{Y}_2,\ldots, \hat{Y}_K)$. These observations are used to initialize the counters ($N_k$'s) that keep track of the number of times the predictions of each arm were observed, and the disagreement values between predictions of each exit point with the first exit $(X_k)$'s. $P_{1k}=X_k/N_k$ gives the estimate of the disagreement probability $\Pr\{\hat{Y}_1\neq \hat{Y}_k\}$. For each arm index, we assign a UCB index which is based on the estimates of the disagreement probabilities and the confidence bonus (line $10$). In each of the subsequent rounds, an arm with the highest UCB index is chosen, denoted as $I_t$. In round $t$, the sample is exited from exit points $I_t$ and observation from all the exit points, i.e., $(\hat{Y}_1, \hat{Y}_2, \ldots, \hat{Y}_{I_t})$, are observed and the values of $N_k, X_k$,  and $P_{1k}$ for all $k\in [I_t]$ are updated (line $12$-$15$). Notice that UEE-UCB is learning the arm with the highest reward $r_k$ which is equivalent to learning the arm with a lower loss $l_k$.

The following theorem characterizes the regret performance of the UEE-UCB algorithm.

\begin{Thm}
Assume multi-exit DNNs satisfy the SD property on the distribution of the input samples. For any $\alpha>1$, the regret of UEE-UCB after $n$ round is given as 

\begin{equation}
\label{eqn:RegretBound}
R(UEE, n)\leq 8\sum_{i \neq i^*} \frac{\log(n)}{\Delta_i}+ (\pi^2/3 +1)\sum_{i \neq i^*}\Delta_i, 
\end{equation}
where $\Delta_i= \gamma_i+c_i -\gamma_{i^*}-c_{i^*}$.
\end{Thm}
\noindent
The proof follows along the same lines as the classical  UCB1 \cite{ML02_UCB1_Auer} after noting that the regret of UEE in round $t$
is 
\begin{eqnarray*}
r_{i^*}-r_{I_t} &= (\gamma_1+c_1- \gamma_{i^*}-c_{i^*})-(\gamma_{1}+c_1-\gamma_{I_t}- c_{I_t})  \\
&=(\gamma_{I_t}+ c_{I_t})- (\gamma_{i^*}+c_{i^*})= l_{I_t}-l_{i^*}.
\end{eqnarray*}
which is the same as in the regret in round $t$ (see Eqn.~\ref{eqn:Regret}).

One can  provide better bounds than given in \ref{eqn:RegretBound} after taking into account the side observations available in each round, i.e., when exit point $i>1$ is selected, we observe $\hat{Y}_i$ and also the values $(\hat{Y}_1, \hat{Y}_2, \ldots, \hat{Y}_{i-1})$ from the previous exit points. However, the regret will still remain $\mathcal{O}(\log n)$.

\section{Training a multi-exit DNN}
\label{sec:TrainingDNN}
As far as the online algorithm evaluation is concerned, a trained multi-exit neural network is apriori given. In this section, we discuss how we  obtain the specific multi-exit DNN that will be used to evaluate the performance of the UEE-UCB algorithm. We follow \cite{liu2021elasticbert} to train a DNN for binary sentiment classification. We begin with an Elastic BERT-Base \cite{liu2021elasticbert} model which is based on a BERT-Base model consisting of $12$ attention layers. An exit is attached after every attention layer of BERT model and trained on a large text corpus with a joint masked language modeling and sentence order prediction loss function across all exits. Once this training is completed, the exits used (MLM and SOP heads) are discarded. The model backbone that remains with the learned weights is the ElasticBERT-Base model. This model is capable of generating language representations that are better suited to early-exit scenarios. We refer to \cite{liu2021elasticbert} for details of the pre-training procedures.

Once we have such a pre-trained model backbone, in the next phase we attach task-specific exits (for example, classification heads) at select points along the backbone architecture and further fine-tune on SST-2 "train" split data \cite{SST2Dataset}. The $[CLS]$ token is a special token that is used to learn sentence-level representations for sentence-level tasks such as sentiment classification. The output representation of this token after each attention module is connected to the classification head (if we plan to attach an exit after that particular layer). A  sketch of the entire training procedure is depicted in Figure~1. $v_1, v_2, v_3$ represents the token embeddings of a given input sentence. The purpose of a head is to produce a representation that can be compared with the given task label to compute a loss term. For example, in a binary sentiment classification task, the task labels are binary. A preferred loss function is binary cross-entropy loss. Hence, the task of the classification head here would be to transform the $d$-dimensional vector representation of the $[CLS]$ token into a probability score via learnable weights.

We might prefer a set of exits that provide us with the maximum range as well as good resolution in the accuracy-cost tradeoff spectrum. Another possibility is to train the model with all the exits attached and later make the choice regarding the exits to retain.

We use Exit Configuration (EC) to denote which exit points on the trained DNN are retained. EC is a binary vector with 1/0 at the $i^{th}$ index representing presence/ absence of a prediction module (binary classification head) after attention layer $(i+1)$ (index starting from 0). Here we choose 3 exit configurations ( We might also identify a model with its exit configuration later.) :
\begin{itemize}
    \item EC-1: 4 exits$[1, 0, 1, 0, 0, 0, 1, 0, 0, 0, 0, 1 ]$
    \item EC-2: 6 exits$[1, 0, 1, 0, 1, 0, 1, 0, 1, 0, 0, 1 ]$
    \item EC-3: 8 exits$[1, 1, 1, 1, 1, 0, 1, 0, 1, 0, 0, 1 ]$
\end{itemize}

The Elastic BERT model with a selected EC is fine-tuned on the "training" split of SST-2 dataset for 5 epochs as per the hyper-parameter choices followed in ElasticBERT \cite{liu2021elasticbert}. Models are checkpointed after every $50$ step. (For 5 epochs with 8544 samples at a batch size of $32$, there are $1335$ steps in total).  The model with the best average accuracy across all exits is chosen as the final model.

\section{Experimental Validation}
\label{sec:Experiment}
In this section, we evaluate the performance of UEE-UCB on different datasets using the ElasticBERT multi-exit DNN in different configurations. We begin with the details of our experimental setup.

\subsection{Setup}
In order to evaluate the performance of our online algorithm, we need a trained multi-exit DNN along with the cost associated with each exit point and a labeled dataset. The input samples (without labels) will simulate the online streaming of samples. The algorithm will select an exit for each of these samples. The labels are only used during the evaluation phase to calculate the actual error probabilities of exits (required in regret calculation). As detailed in the previous section, we have trained $3$ multi-exit DNN with configurations EC-1, EC-2, and EC-3. All the models were trained on SST-2 "train" split dataset. The datasets used for evaluation are
\begin{itemize}
    \item IMDb  \cite{IMDb} dataset: Like SST-2, IMDb is also a movie-review dataset, but the sample distribution may be different from that of SST-2.
    \item Yelp  \cite{Yelp} dataset: This dataset has reviews across diverse entities such as hotels, repair shops, rentals, etc. 
\end{itemize}
The dataset statistics are mentioned in Table~\ref{tab:freq}. In both datasets, we combine all component splits (train, dev, and split) into a single set.
\begin{table}
  \caption{Dataset Statistics}
  \label{tab:freq}
  \begin{tabular}{cccl}
    \toprule
    Dataset&Train&Dev&Test\\
    \midrule

    SST-2 & 8544& 1101 & 2208\\
    IMDb & 20000& 5000& 25000\\
    Yelp  &560000 & -& 38000\\

  \bottomrule
\end{tabular}
\end{table}

The last part of the problem specification is the cost definition along with a scaling parameter $\lambda$. The scaling factor ensures that the error probabilities and costs are directly comparable. The cost structure might vary depending on the particular application. Here, we experiment with two cost structures (CS),
\begin{itemize}
    \item CS-1: Cost associated with a particular exit is the number of attention modules it utilizes to arrive at the prediction, i.e., if a sample exits at the attention module $i$, then the cost is $i$.  This cost structure assumes that each attention module has an equal cost and cost is directly proportional to the amount of computation used across the attention layer. For every problem instance, we set $\lambda = 1/12$, normalizing the cost for direct comparison with the error probabilities.
    \item CS-2: A slabbed cost structuring is utilized to simulate a mobile-cloud co-inference setup where there could be an abrupt jump in cost/ delay if inference needs to proceed to deeper layers deployed in the cloud. We assign a fixed cost of $0.5$ to all exits that use less than or equal to $7$ attention modules for inference (mobile/ edge exits). The rest of the deeper exits (cloud exits) are assigned a cost of $1$ unit. These costs include the scaling factor and are directly compared with the error probabilities.
\end{itemize}  

Given a particular labeled dataset (either Yelp or IMDb) and a multi-exit DNN (either in EC-1, EC-2, or EC-3), we compute the error probabilities for each of the exits (to facilitate regret computation). Further, for a given cost vector, the exit with the lowest sum of cost and error probability is the optimal exit. 

A random ordering of the input samples in a dataset is fed to the algorithm in an online manner (the algorithm does not have access to the true labels). In each round/time step, the algorithm selects an exit and accumulates regret if its choice is not optimal.  We repeat each experiment $20$ times and plot the expected cumulative regret along with a 95$\%$ confidence interval. We consider two baseline policies for benchmarking our algorithm namely Last-exit and Random-exit. In the Last-exit policy, the last exit is chosen in all rounds. This is similar to traditional neural network inference. In Random-exit the exits are selected randomly in each round. We set $\alpha=1$ for all problem instances. Here, a problem instance is specified by a dataset, trained multi-exit DNN, and a cost structure. The number of rounds in each trial is the number of samples in the dataset but we display results only up to around $2000$ as the UEE-UCB saturates and to better highlight the salient features.
\begin{figure}[!t]
  \centering
    \includegraphics[width=\linewidth]{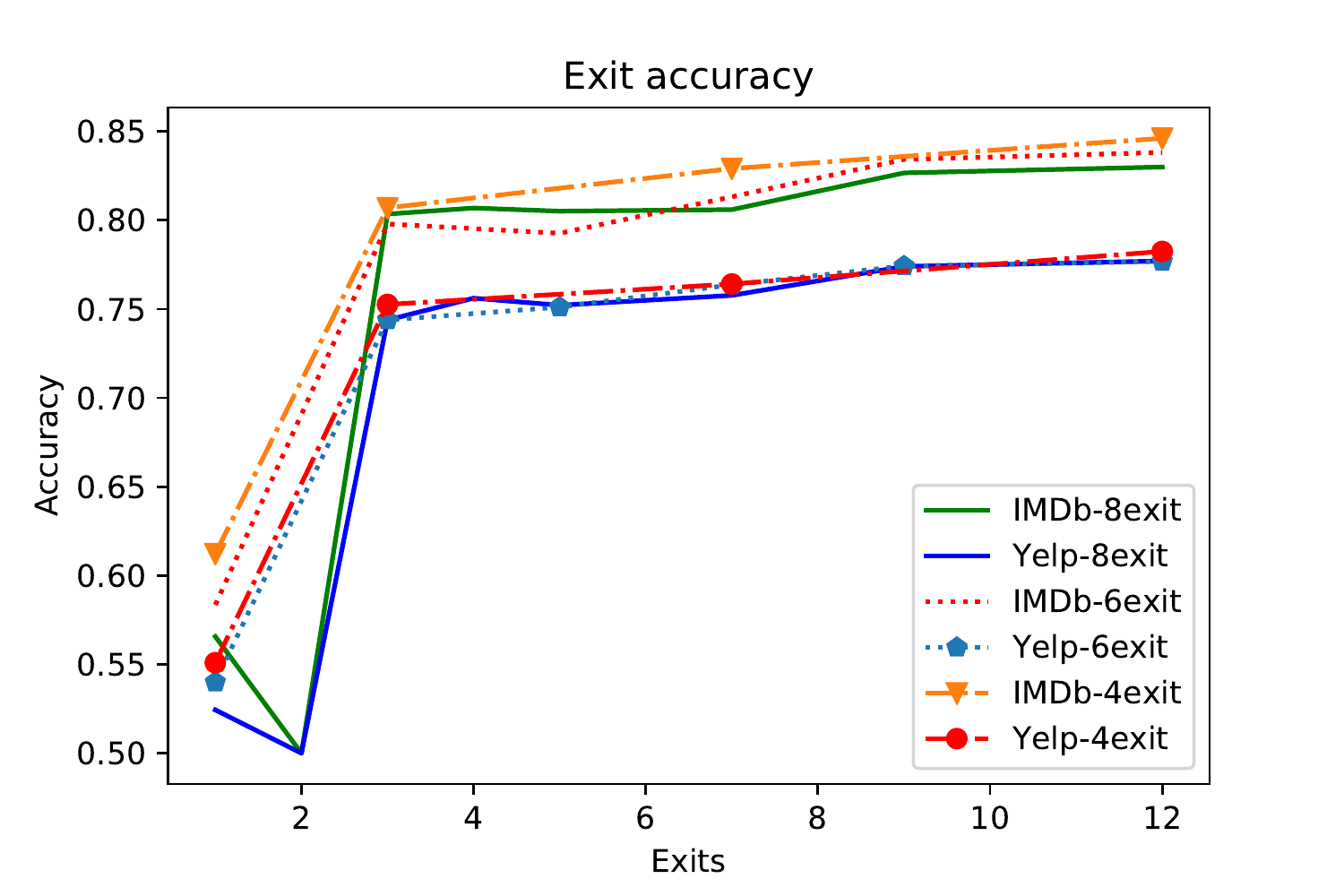}
     \label{fig:ErrorRates}
  \caption{Accuracy vs exits in IMDb and Yelp}
\end{figure}

\subsection{Accuracy of exit points and the SD property}
\begin{table*}
  \caption{Statistics regarding the joint nature of exit predictions}
  \label{tab:commands}
  \begin{tabular}{ccccc|cccc|cccc|cccl}
    \toprule
    Data& 4-exit& 4-exit & 4-exit & 4-exit & 6-exit & 6-exit & 6-exit & 6-exit & 8-exit & 8-exit & 8-exit & 8-exit \\
     set& SD Viol & All Wrg & All Cor & Good & SD Viol & All Wrg & All Cor
     & Good & SD Viol & All Wrg & All Cor
     & Good \\    
    \midrule
    
    IMDB & 14.43$\%$ & 6.59$\%$ &	28.96$\%$	&50.02$\%$	&22.56$\%$	&4.40$\%$	&32.42$\%$	&40.62$\%$ &41.58$\%$	&3.32$\%$	&28.88$\%$	&26.21$\%$ \\
    Yelp & 19.14$\%$ & 8.05$\%$ & 34.08$\%$ &38.73$\%$	&22.46$\%$	&7.21$\%$	&35.05$\%$	&35.28$\%$ &58.35$\%$	&1.56$\%$	&29.99$\%$	&10.09$\%$ \\

    \bottomrule
  \end{tabular}
\end{table*}
Figure~2 shows the accuracy of each exit in models EC-1, EC-2, and EC-3 across datasets, IMDb and Yelp. We observe that adding more exits degrades the accuracy of the final exit. We also observe that as we progress sequentially across the exits, the accuracies get saturated. In both EC-1 (4 exits) as well as EC-2 (6 exits), the accuracies of the exits increase as we go deeper. This is not the case with EC-3 (8-exits). We notice that sufficiently spaced exits have a higher probability of satisfying the assumption of increasing accuracies across exits. This validates our requirement that $\gamma_1 \geq \gamma_2 \geq \ldots \geq \gamma_k$.

We next discuss what fraction of the samples satisfy the SD property. For a better understanding of the samples, we classify them into $4$ groups in each dataset as follows:
\begin{itemize}
    \item Bad samples (SD Viol): For these samples, the exit predictions violate the SD property. This means that for these samples, an early exit made a correct prediction but at least one of the following exits predicts it wrong. All these samples were retained during the algorithm evaluation to check algorithm robustness. We noticed that the share of samples violating the SD property increases with the increase in the number of exit points.
    \item Predictions of all exits are wrong (All Wrg): For input samples in this category, the prediction made by all the exits are the same but they do not match the true label. SD condition is not applicable in this case.
    \item Predictions of all exits are correct (All Cor): For input samples in this category, the prediction made by all the exits are the same and match the true label. For these samples, the SD condition is satisfied.
    \item Good samples (Good): In these samples, predictions flip once while traversing the exits in such a manner that SD property is satisfied. As far as the algorithm is concerned, these samples contain information that discriminates between the exits. It is observed that the number of good samples reduces with an increase in the number of exits.
\end{itemize}

Table~\ref{tab:commands} gives details of the fraction of samples in each category for each of the datasets and different configurations of multi-exit DNNs. As seen the fraction of samples that violates the SD condition increase with the number of exits. Also, this violation is more in the Yelp datasets than in the IMDb dataset. 

Figure~3 highlights the effect of samples that violate SD property in EC-3 on the Yelp dataset with CS-2. Yelp dataset with EC-3 and CS-2 configuration is the hardest among all configurations as there are more exits and loss values of the exits points are close to each other. Apart from the full dataset setting (with all 598000 samples), we consider settings with $50\%$, $75\%$, and $100\%$ of the bad samples removed. The regret plot shows that a reduction in the number of samples that violate the SD property results in better algorithm performance. Even when the violation is $25\%$ the algorithm is able to achieve sublinear regret. Setting aside these cases, it is observed that the algorithm is robust to around 20$\%$ of bad samples in the considered settings. Interestingly, it achieves sub-linear regret for the configuration with 8 exits (EC-3) model on the IMDb dataset, where the share of bad samples is around 40$\%$ (see Figures 4 \& 5). Amongst other factors, the performance of the algorithm also depends on the specific cost vector employed. While experimenting, we noted that if the exits are sufficiently spaced, then assumptions hold to a great degree and the algorithm performs well.

\begin{figure}[h]
  \centering
  \includegraphics[width=\linewidth]{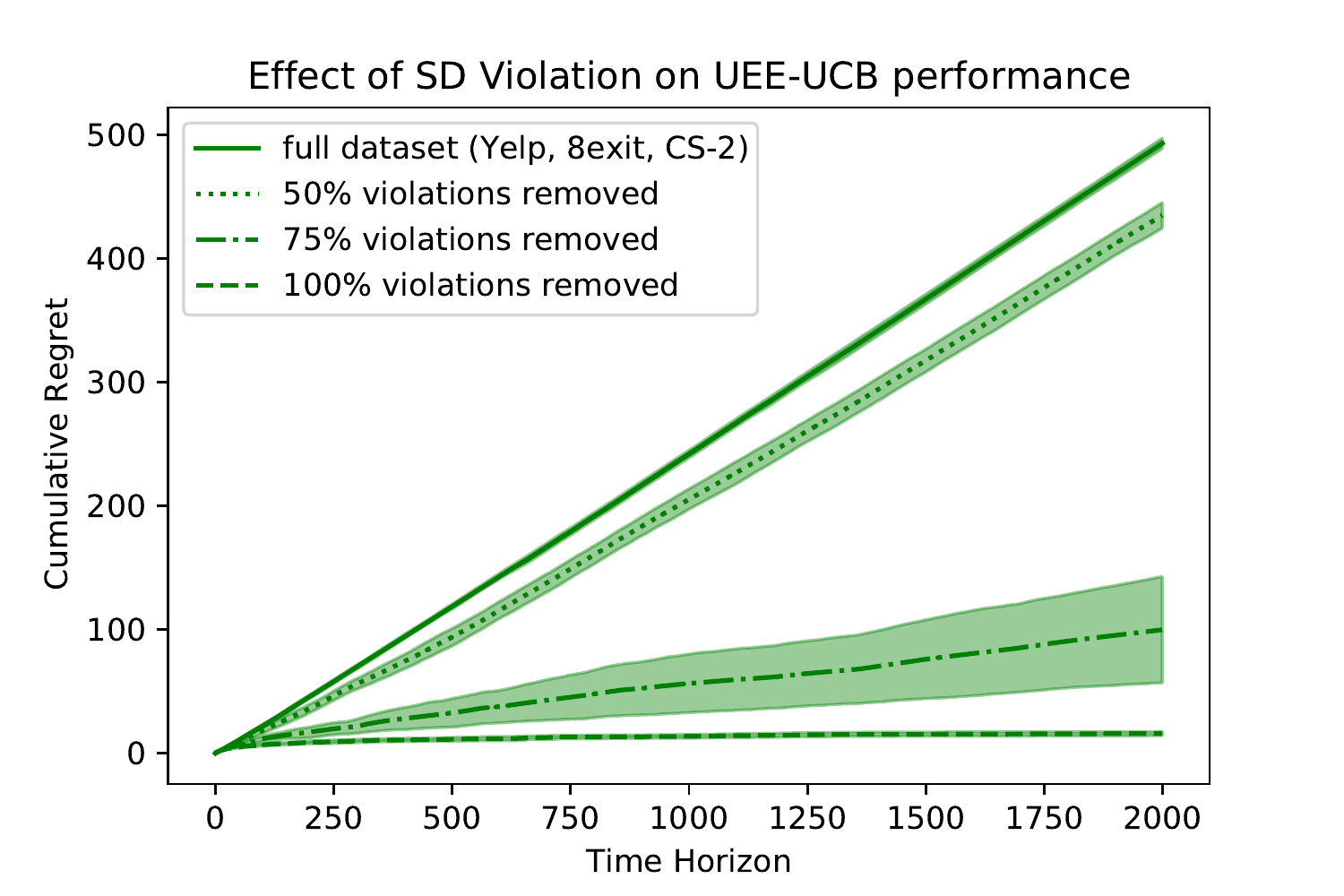}
  
  \caption{Cumulative Regret Curves highlighting the effect of SD Violation on algorithm performance, Expected Cumulative Regret vs Time (Number of samples processed)}
  \Description{Cumulative Regret plot for 4-exit MEDNN IMDb dataset}
\end{figure}

\begin{table*}
  \caption{Accuracy-Cost Trade-Offs }
  \label{tab:freq}
  \begin{tabular}{ccc|cc|cc|cc|cc|ccl}
    \toprule
    Policy &4-exit&IMDb&4-exit&Yelp&6-exit&IMDb&6-exit&Yelp&8-exit&IMDb&8-exit&Yelp\\    \hline
     &Acc&Cost&Acc&Cost&Acc&Cost&Acc&Cost&Acc&Cost&Acc&Cost\\
    \midrule

    UEE-UCB & 80.71& 0.25$x$& 75.26& 0.25$x$& 79.79& 0.25$x$& 74.39 & 0.25$x$& 80.10& 0.25$x$& 50.00 & 0.17$x$  \\
    Last-Exit & 84.61& 1$x$& 78.24& 1$x$& 83.82& 1$x$& 77.68& 1$x$& 83.00& 1$x$& 77.71& 1$x$\\
    Random  &77.34 & 0.48$x$&71.24 & 0.48$x$&77.65 & 0.51$x$ &72.50 & 0.51$x$&74.22 & 0.48$x$ &69.82 & 0.45$x$\\

  \bottomrule
\end{tabular}
\end{table*}

\subsection{Regret Performance}

The expected cumulative regret curves for the proposed problem settings are plotted in Figures~4 \& 5. Please note that in all the regret curves, the time horizon represents the number of samples that the algorithm has accessed so far ( not the actual time taken).  We observe that UEE-UCB beats the baselines and achieves sub-linear regret in all but one case. The outlier case is for the configuration with 8 exits (EC-3) on the Yelp dataset. This is expected since from Table~\ref{tab:commands} it is clear that more than 50$\%$ of the samples violate SD property. 

Table~\ref{tab:freq} reports the accuracy-cost trade-off that can be achieved by UEE-UCB when compared to the baseline algorithms. The experiments were conducted on the CS-1 cost structure. The results are averaged over 20 trials. In each trial, randomly shuffled data is fed in an online manner, at each round, the policy chooses an exit. Accuracy is computed considering the chosen exit prediction. The costs reported are the per-sample averaged costs. For example, if we take the 4-exit IMDb case, we can get a cost reduction of 75$\%$ for an accuracy drop of around 4$\%$ with respect to the last exit policy. A significant reduction in the cost for a comparatively minor loss in accuracy is observed for most of the other settings compared to the baselines.

A possible process pipeline that could be adopted to deploy our method is as follows:  Once we have a multi-exit DNN model trained on a specific dataset and have a target data distribution over which the model needs to be deployed, we follow two steps: 1) collect a representative subset of labeled data from the target distribution This can be used to validate whether the assumptions required by our online learning algorithm hold on the target distribution. 2) if the assumptions are validated,  UEE-UCB can be launched in the unsupervised online setting to learn the optimal exit point.

\begin{figure}[h]
  \centering
  \includegraphics[width=\linewidth]{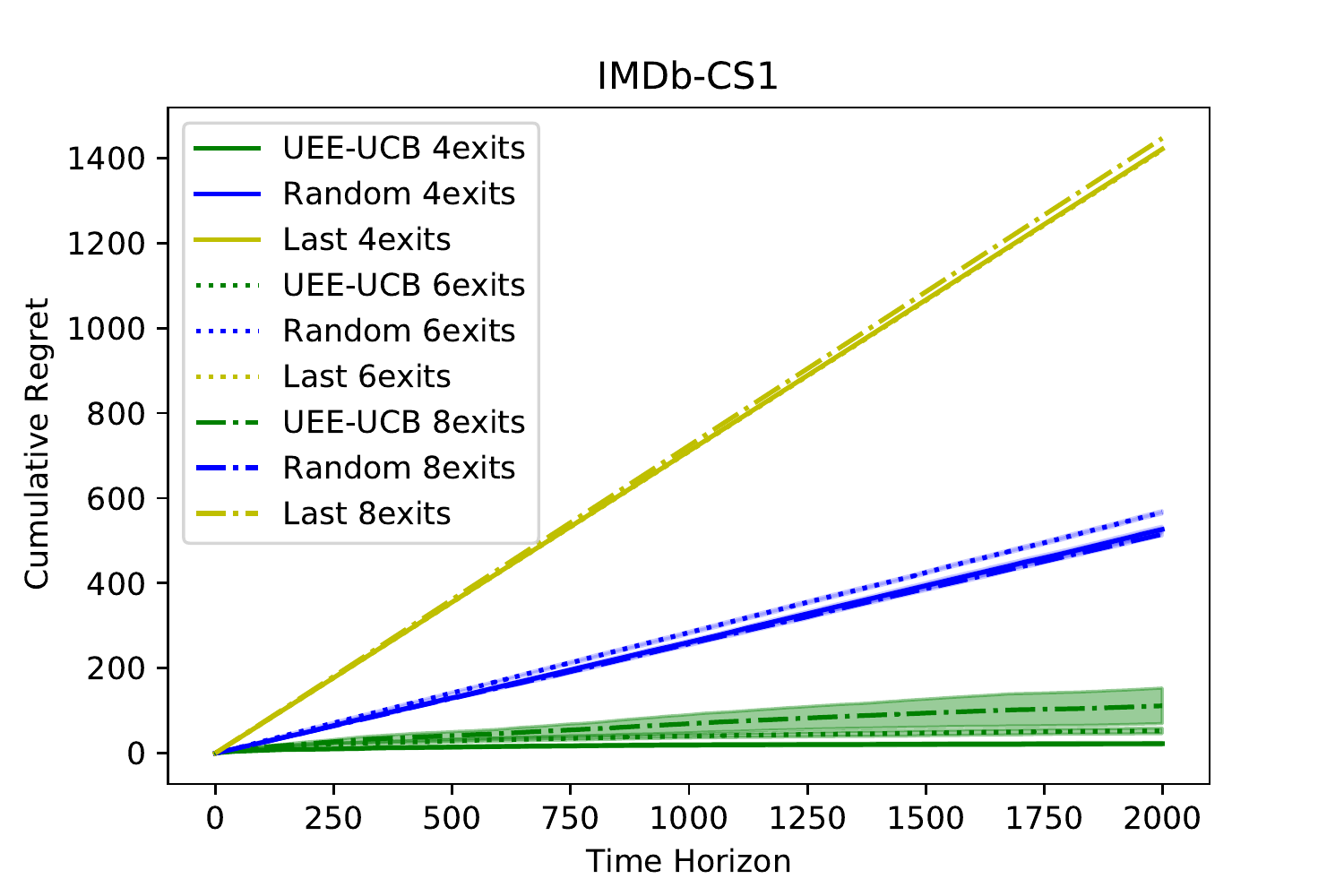}
  \includegraphics[width=\linewidth]{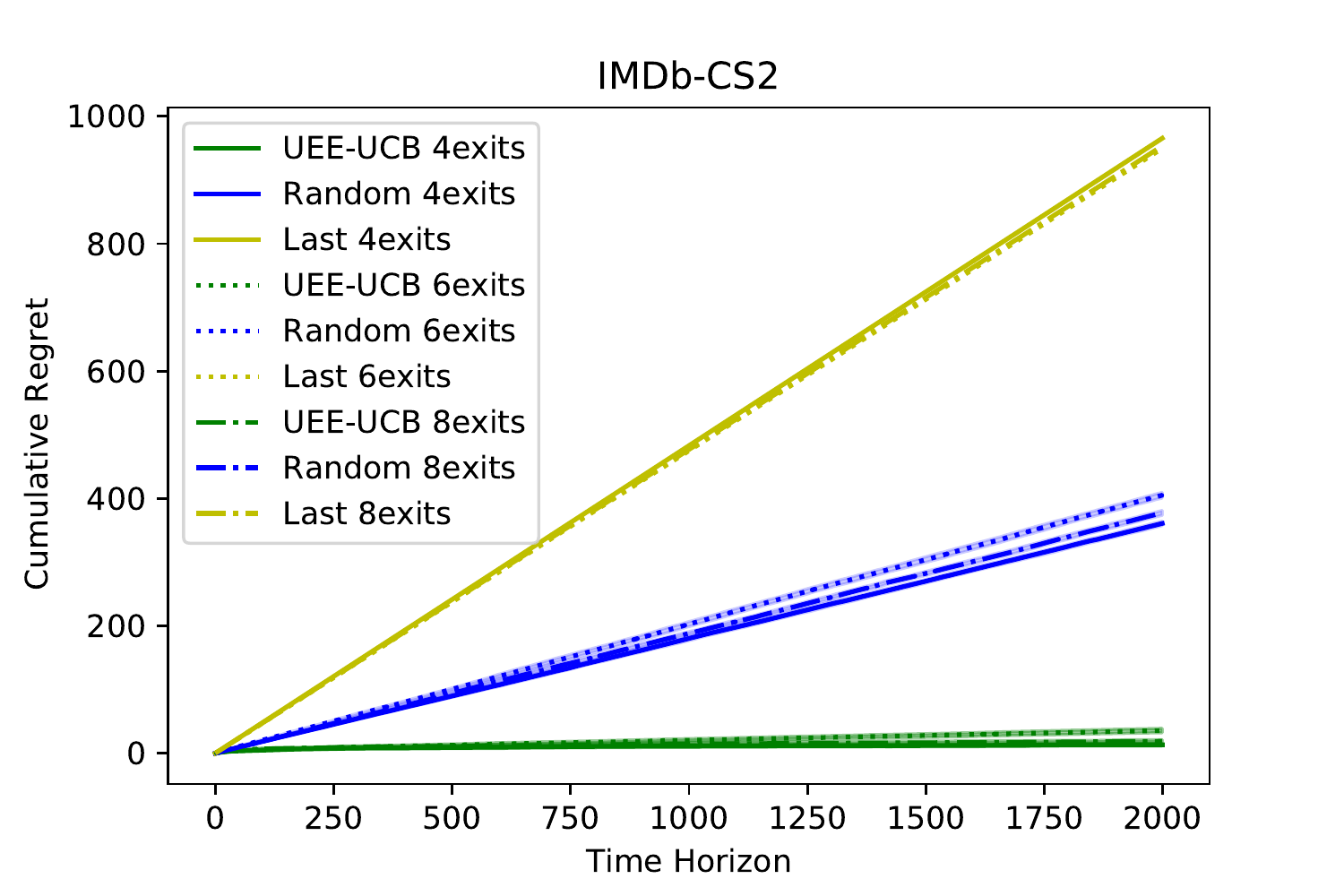}
  
  \caption{Expected Cumulative Regret vs Time ( Number of samples processed) plot for UEE-UCB and baselines tested on IMDb with Cost Structures, CS-1 and CS-2 }
  \Description{Cumulative Regret plot for 4-exit MEDNN IMDb dataset}
\end{figure}

\begin{figure}[h]
  \centering
  \includegraphics[width=\linewidth]{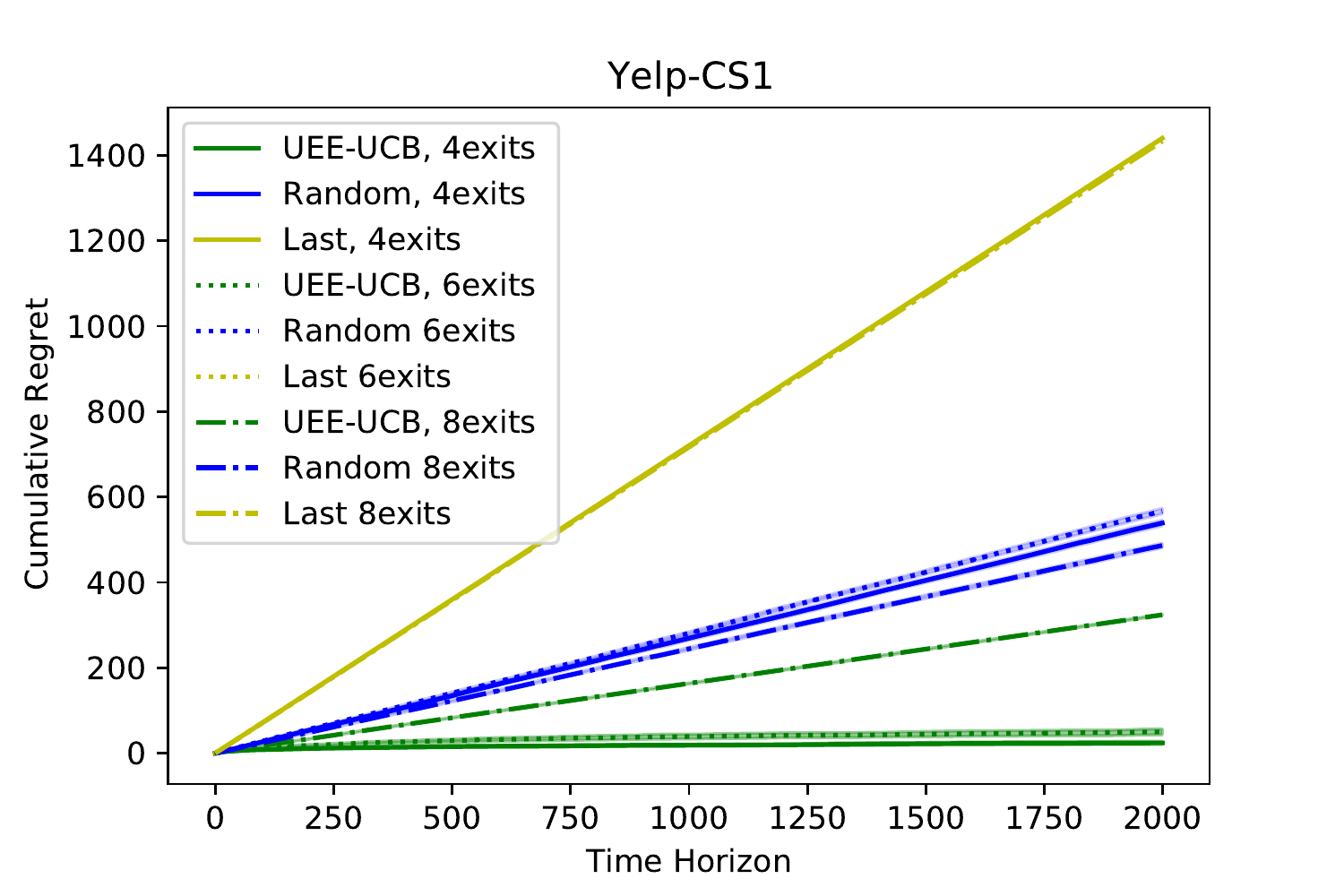}
  \includegraphics[width=\linewidth]{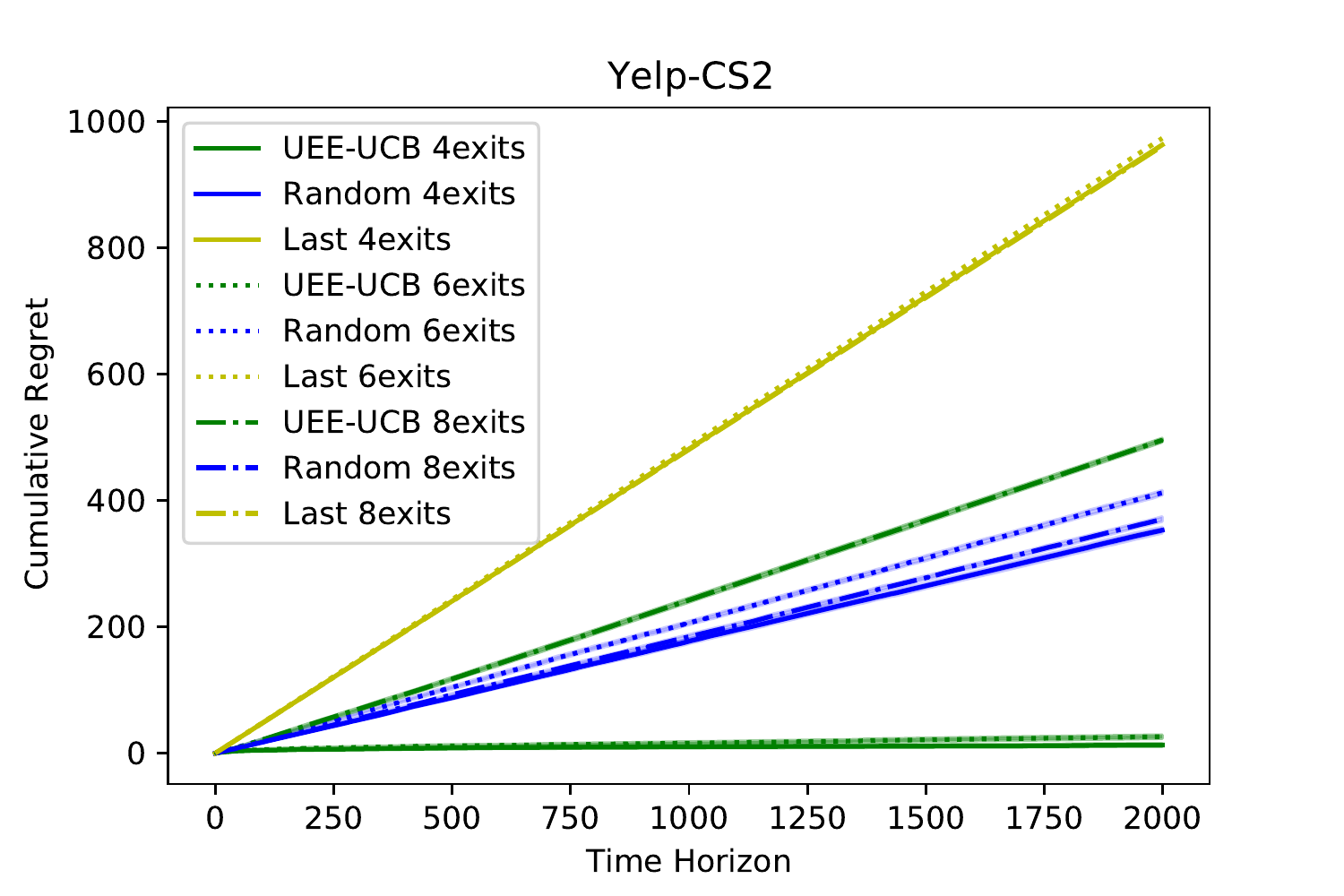}
  \caption{Expected Cumulative Regret vs Time ( Number of samples processed) Plot for UEE-UCB and baselines tested on Yelp with Cost Structures, CS-1 and CS-2 }
  \Description{A woman and a girl in white dresses sit in an open car.}
\end{figure}

\section{Conclusion and Future Work}
\label{sec:Conclusion}
The size of neural networks used in various applications is growing with more layers added to achieve higher accuracy. However, this higher accuracy comes at the cost of more latency or requires more computational resources. Often, all the samples input to the DNNs are not of the same difficulty level, and hence all need not pass through all the layers of the DNNs. This has spurred the research of multi-exit in DNNs, where DNNs are attached with inference blocks at the intermediate  layers instead of only at the last layer giving rise to multi-exit DNNs. However, the challenge in these multi-exit DNNs is to design criteria to decide whether to exit or continue to process at each exit point or to decide which  exit points give the best accuracy and cost trade-off. The current multi-exit DNNs use a fixed threshold at each exit point against which entropy scores are compared to make a decision on whether to continue or exit. However, the problem with a fixed threshold is that it is not domain-specific and the same fixed threshold may not work for all the domains. Thus, there is a need for mechanisms to learn the optimal exit that is domain-specific.

In this work, we developed an online learning algorithm that learns the exit points that gives the best trade-off between accuracy and cost (latency, computational resource) as it gets to see the samples from the domain.  The main feature of our algorithm, named Unsupervised Early Exit (UEE) is that it does not need to know the ground truth labels at any point and hence is entirely unsupervised. The algorithm exploits the Strong Dominance (SD) property that the multi-exit DNNs are expected to satisfy due to their inherent design. Specifically, we focused on the ElasticBERT multi-exit DNNs and demonstrated that it satisfies the SD property on the IMDb and Yelp datasets. Experiments on these datasets demonstrated that UEE quickly learns the optimal exit resulting in small regret compared to the benchmark policies.

Our work focused on developing an algorithm that identifies an exit point with the best accuracy-cost across all the samples as is the case in most online learning setups. Also, the UEE-UCB algorithm a priori decides where to exit without looking into the outcome of the intermediate exit point.  However, the optimal exit point can change depending on the sample which could be interpreted as a context. One interesting future direction is to extend our unsupervised online learning setting to the case where exit points for each sample are decided based on the outcome of exit points that the sample passes through.

We studied  ElasticBERT drawn from the application of natural language processing and verified that it largely follows SD property on IMDb and Yelp datasets. Multi-exit DNNs are also used in other domains like image processing and ranking systems. One can verify if they satisfy the SD property on the datasets like ImageNet and evaluate the performance of the UEE-UCB algorithm. Another interesting direction is to look for structural properties other than SD that can help to learn the optimal exit in an unsupervised setting. 

\section*{Acknowledgements}
Manjesh K. Hanwal acknowledges funding from SERB under the MATRICS grant (MTR/2021/000645).

\bibliographystyle{ACM-Reference-Format}
\bibliography{Biblio}


\begin{thebibliography}{29}


\ifx \showCODEN    \undefined \def \showCODEN     #1{\unskip}     \fi
\ifx \showDOI      \undefined \def \showDOI       #1{#1}\fi
\ifx \showISBNx    \undefined \def \showISBNx     #1{\unskip}     \fi
\ifx \showISBNxiii \undefined \def \showISBNxiii  #1{\unskip}     \fi
\ifx \showISSN     \undefined \def \showISSN      #1{\unskip}     \fi
\ifx \showLCCN     \undefined \def \showLCCN      #1{\unskip}     \fi
\ifx \shownote     \undefined \def \shownote      #1{#1}          \fi
\ifx \showarticletitle \undefined \def \showarticletitle #1{#1}   \fi
\ifx \showURL      \undefined \def \showURL       {\relax}        \fi
\providecommand\bibfield[2]{#2}
\providecommand\bibinfo[2]{#2}
\providecommand\natexlab[1]{#1}
\providecommand\showeprint[2][]{arXiv:#2}

\bibitem[Asghar(2016)]%
        {Yelp}
\bibfield{author}{\bibinfo{person}{Nabiha Asghar}.}
  \bibinfo{year}{2016}\natexlab{}.
\newblock \showarticletitle{Yelp Dataset Challenge: Review Rating Prediction}.
\newblock \bibinfo{journal}{\emph{CoRR}}  \bibinfo{volume}{abs/1605.05362}
  (\bibinfo{year}{2016}).
\newblock
\urldef\tempurl%
\url{http://arxiv.org/abs/1605.05362}
\showURL{%
\tempurl}


\bibitem[Auer et~al\mbox{.}(2002)]%
        {ML02_UCB1_Auer}
\bibfield{author}{\bibinfo{person}{Peter Auer} {et~al\mbox{.}}}
  \bibinfo{year}{2002}\natexlab{}.
\newblock \showarticletitle{Finite-time Analysis of the Multiarmed Bandit
  Problem}.
\newblock \bibinfo{journal}{\emph{Machine Learning}}  \bibinfo{volume}{47}
  (\bibinfo{year}{2002}), \bibinfo{pages}{235--256}.
\newblock


\bibitem[Bapna et~al\mbox{.}(2020)]%
        {bapna2020controlling}
\bibfield{author}{\bibinfo{person}{Ankur Bapna}, \bibinfo{person}{Naveen
  Arivazhagan}, {and} \bibinfo{person}{Orhan Firat}.}
  \bibinfo{year}{2020}\natexlab{}.
\newblock \showarticletitle{Controlling computation versus quality for neural
  sequence models}.
\newblock \bibinfo{journal}{\emph{arXiv preprint arXiv:2002.07106}}
  (\bibinfo{year}{2020}).
\newblock


\bibitem[Bartók et~al\mbox{.}(2014)]%
        {partialmonitoring}
\bibfield{author}{\bibinfo{person}{Gábor Bartók}, \bibinfo{person}{Dean~P.
  Foster}, \bibinfo{person}{Dávid Pál}, \bibinfo{person}{Alexander Rakhlin},
  {and} \bibinfo{person}{Csaba Szepesvári}.} \bibinfo{year}{2014}\natexlab{}.
\newblock \showarticletitle{Partial Monitoring—Classification, Regret Bounds,
  and Algorithms}.
\newblock \bibinfo{journal}{\emph{Mathematics of Operations Research}}
  \bibinfo{volume}{39}, \bibinfo{number}{4} (\bibinfo{year}{2014}),
  \bibinfo{pages}{967--997}.
\newblock


\bibitem[Cambazoglu et~al\mbox{.}(2010)]%
        {WSDM2010_EarlyExitRanking}
\bibfield{author}{\bibinfo{person}{B.~Barla Cambazoglu}, \bibinfo{person}{Hugo
  Zaragoza}, \bibinfo{person}{Olivier Chapelle}, \bibinfo{person}{Jiang Chen},
  \bibinfo{person}{Ciya Liao}, \bibinfo{person}{Zhaohui Zheng}, {and}
  \bibinfo{person}{Jon Degenhardt}.} \bibinfo{year}{2010}\natexlab{}.
\newblock \showarticletitle{Early exit optimizations for additive machine
  learned ranking systems}. In \bibinfo{booktitle}{\emph{Proceedings of the
  Third ACM International Conference on Web Search and Data Mining (WSDM}}.
  \bibinfo{pages}{411--420}.
\newblock


\bibitem[Dabre et~al\mbox{.}(2020)]%
        {DBLP:conf/aclnmt/DabreRF20}
\bibfield{author}{\bibinfo{person}{Raj Dabre}, \bibinfo{person}{Raphael
  Rubino}, {and} \bibinfo{person}{Atsushi Fujita}.}
  \bibinfo{year}{2020}\natexlab{}.
\newblock \showarticletitle{Balancing Cost and Benefit with Tied-Multi
  Transformers}. In \bibinfo{booktitle}{\emph{Proceedings of the Fourth
  Workshop on Neural Generation and Translation, NGT@ACL 2020, Online, July
  5-10, 2020}}, \bibfield{editor}{\bibinfo{person}{Alexandra Birch},
  \bibinfo{person}{Andrew~M. Finch}, \bibinfo{person}{Hiroaki Hayashi},
  \bibinfo{person}{Kenneth Heafield}, \bibinfo{person}{Marcin
  Junczys{-}Dowmunt}, \bibinfo{person}{Ioannis Konstas}, \bibinfo{person}{Xian
  Li}, \bibinfo{person}{Graham Neubig}, {and} \bibinfo{person}{Yusuke Oda}}
  (Eds.). \bibinfo{publisher}{Association for Computational Linguistics},
  \bibinfo{pages}{24--34}.
\newblock
\urldef\tempurl%
\url{https://doi.org/10.18653/v1/2020.ngt-1.3}
\showDOI{\tempurl}


\bibitem[Dai et~al\mbox{.}(2020)]%
        {dai2020epnet}
\bibfield{author}{\bibinfo{person}{Xin Dai}, \bibinfo{person}{Xiangnan Kong},
  {and} \bibinfo{person}{Tian Guo}.} \bibinfo{year}{2020}\natexlab{}.
\newblock \showarticletitle{{EPNet}: Learning to exit with flexible
  multi-branch network}. In \bibinfo{booktitle}{\emph{ACM Int. Conf. on
  Information \& Knowledge Management}}. \bibinfo{pages}{235--244}.
\newblock


\bibitem[Devlin et~al\mbox{.}(2019)]%
        {BERTPretrain}
\bibfield{author}{\bibinfo{person}{Jacob Devlin}, \bibinfo{person}{Ming-Wei
  Chang}, \bibinfo{person}{Kenton Lee}, {and} \bibinfo{person}{Kristina
  Toutanova}.} \bibinfo{year}{2019}\natexlab{}.
\newblock \showarticletitle{BERT: Pre-training of deep bidirectional
  transformers for language understanding}. In \bibinfo{booktitle}{\emph{In
  Proceedings of the 2019 Conference of the North American Chapter of the
  Association for Computational Linguistics: Human Language Technologies,
  Volume 1}}. \bibinfo{pages}{4171-- 4186}.
\newblock


\bibitem[Elbayad et~al\mbox{.}(2020)]%
        {elbayad19arxiv}
\bibfield{author}{\bibinfo{person}{Maha Elbayad}, \bibinfo{person}{Jiatao Gu},
  \bibinfo{person}{Edouard Grave}, {and} \bibinfo{person}{Michael Auli}.}
  \bibinfo{year}{2020}\natexlab{}.
\newblock \showarticletitle{Depth-Adaptive Transformer}. In
  \bibinfo{booktitle}{\emph{In Proc. of ICLR}}.
\newblock


\bibitem[Fang et~al\mbox{.}(2020)]%
        {fang2020flexdnn}
\bibfield{author}{\bibinfo{person}{Biyi Fang}, \bibinfo{person}{Xiao Zeng},
  \bibinfo{person}{Faen Zhang}, \bibinfo{person}{Hui Xu}, {and}
  \bibinfo{person}{Mi Zhang}.} \bibinfo{year}{2020}\natexlab{}.
\newblock \showarticletitle{FlexDNN: Input-adaptive on-device deep learning for
  efficient mobile vision}. In \bibinfo{booktitle}{\emph{IEEE/ACM Symposium on
  Edge Computing (SEC)}}. \bibinfo{pages}{84--95}.
\newblock


\bibitem[Hanawal et~al\mbox{.}(2017)]%
        {hanawal2017unsupervised}
\bibfield{author}{\bibinfo{person}{Manjesh Hanawal}, \bibinfo{person}{Csaba
  Szepesvari}, {and} \bibinfo{person}{Venkatesh Saligrama}.}
  \bibinfo{year}{2017}\natexlab{}.
\newblock \showarticletitle{Unsupervised sequential sensor acquisition}. In
  \bibinfo{booktitle}{\emph{Artificial Intelligence and Statistics}}. PMLR,
  \bibinfo{pages}{803--811}.
\newblock


\bibitem[Huang et~al\mbox{.}(2018)]%
        {ICLR2018_ResourceEfficient}
\bibfield{author}{\bibinfo{person}{JGao Huang}, \bibinfo{person}{Danlu Chen},
  \bibinfo{person}{Tianhong Li}, \bibinfo{person}{Felix Wu},
  \bibinfo{person}{Laurens van~der Maaten}, {and} \bibinfo{person}{Kilian
  Weinberger}.} \bibinfo{year}{2018}\natexlab{}.
\newblock \showarticletitle{Multi-scale dense networks for resource efficient
  image classification}. In \bibinfo{booktitle}{\emph{In Proceedings of the 6th
  International Conference on Learning Representations}}.
\newblock


\bibitem[Ju et~al\mbox{.}(2021a)]%
        {ju2021dynamic}
\bibfield{author}{\bibinfo{person}{Weiyu Ju}, \bibinfo{person}{Wei Bao},
  \bibinfo{person}{Liming Ge}, {and} \bibinfo{person}{Dong Yuan}.}
  \bibinfo{year}{2021}\natexlab{a}.
\newblock \showarticletitle{Dynamic Early Exit Scheduling for Deep Neural
  Network Inference through Contextual Bandits}. In
  \bibinfo{booktitle}{\emph{ACM Int. Conf. on Information \& Knowledge
  Management}}. \bibinfo{pages}{823--832}.
\newblock


\bibitem[Ju et~al\mbox{.}(2021b)]%
        {ju2021learning}
\bibfield{author}{\bibinfo{person}{Weiyu Ju}, \bibinfo{person}{Wei Bao},
  \bibinfo{person}{Dong Yuan}, \bibinfo{person}{Liming Ge}, {and}
  \bibinfo{person}{Bing~Bing Zhou}.} \bibinfo{year}{2021}\natexlab{b}.
\newblock \showarticletitle{Learning Early Exit for Deep Neural Network
  Inference on Mobile Devices through Multi-Armed Bandits}. In
  \bibinfo{booktitle}{\emph{IEEE/ACM Int. Symposium on Cluster, Cloud and
  Internet Computing (CCGrid)}}. \bibinfo{pages}{11--20}.
\newblock


\bibitem[Kaya et~al\mbox{.}(2019)]%
        {kaya2019shallow}
\bibfield{author}{\bibinfo{person}{Yigitcan Kaya}, \bibinfo{person}{Sanghyun
  Hong}, {and} \bibinfo{person}{Tudor Dumitras}.}
  \bibinfo{year}{2019}\natexlab{}.
\newblock \showarticletitle{Shallow-deep networks: Understanding and mitigating
  network overthinking}. In \bibinfo{booktitle}{\emph{International conference
  on machine learning}}. PMLR, \bibinfo{pages}{3301--3310}.
\newblock


\bibitem[Kim and Park(2020)]%
        {kim2020low}
\bibfield{author}{\bibinfo{person}{Geonho Kim} {and} \bibinfo{person}{Jongsun
  Park}.} \bibinfo{year}{2020}\natexlab{}.
\newblock \showarticletitle{Low Cost Early Exit Decision Unit Design for CNN
  Accelerator}. In \bibinfo{booktitle}{\emph{IEEE Int. SoC Design Conf.}}
  \bibinfo{pages}{127--128}.
\newblock


\bibitem[Laskaridis et~al\mbox{.}(2020)]%
        {laskaridis2020spinn}
\bibfield{author}{\bibinfo{person}{Stefanos Laskaridis},
  \bibinfo{person}{Stylianos~I Venieris}, \bibinfo{person}{Mario Almeida},
  \bibinfo{person}{Ilias Leontiadis}, {and} \bibinfo{person}{Nicholas~D Lane}.}
  \bibinfo{year}{2020}\natexlab{}.
\newblock \showarticletitle{{SPINN}: synergistic progressive inference of
  neural networks over device and cloud}. In \bibinfo{booktitle}{\emph{Int.
  Conf. on Mobile Computing and Networking (MobiCom)}}. \bibinfo{pages}{1--15}.
\newblock


\bibitem[Li et~al\mbox{.}(2019)]%
        {li2019edge}
\bibfield{author}{\bibinfo{person}{En Li}, \bibinfo{person}{Liekang Zeng},
  \bibinfo{person}{Zhi Zhou}, {and} \bibinfo{person}{Xu Chen}.}
  \bibinfo{year}{2019}\natexlab{}.
\newblock \showarticletitle{Edge AI: On-demand accelerating deep neural network
  inference via edge computing}.
\newblock \bibinfo{journal}{\emph{IEEE Transactions on Wireless
  Communications}} \bibinfo{volume}{19}, \bibinfo{number}{1}
  (\bibinfo{year}{2019}), \bibinfo{pages}{447--457}.
\newblock


\bibitem[Liu et~al\mbox{.}(2021)]%
        {liu2021elasticbert}
\bibfield{author}{\bibinfo{person}{Xiangyang Liu}, \bibinfo{person}{Tianxiang
  Sun}, \bibinfo{person}{Junliang He}, \bibinfo{person}{Lingling Wu},
  \bibinfo{person}{Xinyu Zhang}, \bibinfo{person}{Hao Jiang},
  \bibinfo{person}{Zhao Cao}, \bibinfo{person}{Xuanjing Huang}, {and}
  \bibinfo{person}{Xipeng Qiu}.} \bibinfo{year}{2021}\natexlab{}.
\newblock \showarticletitle{Towards Efficient {NLP:} {A} Standard Evaluation
  and {A} Strong Baseline}.
\newblock  (\bibinfo{year}{2021}).
\newblock
\urldef\tempurl%
\url{https://arxiv.org/abs/2110.07038}
\showURL{%
\tempurl}


\bibitem[Maas et~al\mbox{.}(2011)]%
        {IMDb}
\bibfield{author}{\bibinfo{person}{Andrew~L. Maas}, \bibinfo{person}{Raymond~E.
  Daly}, \bibinfo{person}{Peter~T. Pham}, \bibinfo{person}{Dan Huang},
  \bibinfo{person}{Andrew~Y. Ng}, {and} \bibinfo{person}{Christopher Potts}.}
  \bibinfo{year}{2011}\natexlab{}.
\newblock \showarticletitle{Learning Word Vectors for Sentiment Analysis}. In
  \bibinfo{booktitle}{\emph{Proceedings of the 49th Annual Meeting of the
  Association for Computational Linguistics: Human Language Technologies}}.
  \bibinfo{publisher}{Association for Computational Linguistics},
  \bibinfo{address}{Portland, Oregon, USA}, \bibinfo{pages}{142--150}.
\newblock
\urldef\tempurl%
\url{http://www.aclweb.org/anthology/P11-1015}
\showURL{%
\tempurl}


\bibitem[Pacheco et~al\mbox{.}(2021a)]%
        {pacheco2021calibration}
\bibfield{author}{\bibinfo{person}{Roberto~G. Pacheco},
  \bibinfo{person}{Rodrigo~S. Couto}, {and} \bibinfo{person}{O. Simeone}.}
  \bibinfo{year}{2021}\natexlab{a}.
\newblock \showarticletitle{Calibration-Aided Edge Inference Offloading via
  Adaptive Model Partitioning of Deep Neural Networks}. In
  \bibinfo{booktitle}{\emph{IEEE Int. Conf. on Communications (ICC)}}.
  \bibinfo{pages}{1--6}.
\newblock


\bibitem[Pacheco et~al\mbox{.}(2021b)]%
        {pacheco2021distorted}
\bibfield{author}{\bibinfo{person}{Roberto~G. Pacheco},
  \bibinfo{person}{Fernanda D. V.~R. Oliveira}, {and}
  \bibinfo{person}{Rodrigo~S. Couto}.} \bibinfo{year}{2021}\natexlab{b}.
\newblock \showarticletitle{Early-exit deep neural networks for distorted
  images: providing an efficient edge offloading}. In
  \bibinfo{booktitle}{\emph{IEEE Global Communications Conf. (GLOBECOM)}}.
  \bibinfo{pages}{1--6}.
\newblock
\urldef\tempurl%
\url{https://doi.org/10.1109/GLOBECOM46510.2021.9685469}
\showDOI{\tempurl}


\bibitem[Scardapane et~al\mbox{.}(2020)]%
        {Scardapane_2020}
\bibfield{author}{\bibinfo{person}{Simone Scardapane}, \bibinfo{person}{Michele
  Scarpiniti}, \bibinfo{person}{Enzo Baccarelli}, {and}
  \bibinfo{person}{Aurelio Uncini}.} \bibinfo{year}{2020}\natexlab{}.
\newblock \showarticletitle{Why Should We Add Early Exits to Neural Networks?}
\newblock \bibinfo{journal}{\emph{Cognitive Computation}} \bibinfo{volume}{12},
  \bibinfo{number}{5} (\bibinfo{date}{jun} \bibinfo{year}{2020}),
  \bibinfo{pages}{954--966}.
\newblock
\urldef\tempurl%
\url{https://doi.org/10.1007/s12559-020-09734-4}
\showDOI{\tempurl}


\bibitem[Socher et~al\mbox{.}(2013)]%
        {SST2Dataset}
\bibfield{author}{\bibinfo{person}{Richard Socher}, \bibinfo{person}{Alex
  Perelygin}, \bibinfo{person}{Jean Wu}, \bibinfo{person}{Jason Chuang},
  \bibinfo{person}{Christopher~D. Manning}, \bibinfo{person}{Andrew Ng}, {and}
  \bibinfo{person}{Christopher Potts}.} \bibinfo{year}{2013}\natexlab{}.
\newblock \showarticletitle{Recursive Deep Models for Semantic Compositionality
  Over a Sentiment Treebank}. In \bibinfo{booktitle}{\emph{Proceedings of the
  2013 Conference on Empirical Methods in Natural Language Processing}}.
  \bibinfo{pages}{1631--1642}.
\newblock


\bibitem[Teerapittayanon et~al\mbox{.}(2016)]%
        {teerapittayanon2016branchynet}
\bibfield{author}{\bibinfo{person}{Surat Teerapittayanon},
  \bibinfo{person}{Bradley McDanel}, {and} \bibinfo{person}{Hsiang-Tsung
  Kung}.} \bibinfo{year}{2016}\natexlab{}.
\newblock \showarticletitle{Branchynet: Fast inference via early exiting from
  deep neural networks}. In \bibinfo{booktitle}{\emph{2016 23rd International
  Conference on Pattern Recognition (ICPR)}}. IEEE,
  \bibinfo{pages}{2464--2469}.
\newblock


\bibitem[Verma et~al\mbox{.}(2019)]%
        {verma2019Unsupervised}
\bibfield{author}{\bibinfo{person}{Arun Verma}, \bibinfo{person}{Manjesh
  Hanawal}, \bibinfo{person}{Csaba Szepesvari}, {and}
  \bibinfo{person}{Venkatesh Saligrama}.} \bibinfo{year}{2019}\natexlab{}.
\newblock \showarticletitle{Online Algorithm for Unsupervised Sensor
  Selection}. In \bibinfo{booktitle}{\emph{Proceedings of the Twenty-Second
  International Conference on Artificial Intelligence and Statistics}}. PMLR,
  \bibinfo{pages}{3168--3176}.
\newblock


\bibitem[Wang et~al\mbox{.}(2019b)]%
        {wang2019dynexit}
\bibfield{author}{\bibinfo{person}{Meiqi Wang}, \bibinfo{person}{Jianqiao Mo},
  \bibinfo{person}{Jun Lin}, \bibinfo{person}{Zhongfeng Wang}, {and}
  \bibinfo{person}{Li Du}.} \bibinfo{year}{2019}\natexlab{b}.
\newblock \showarticletitle{DynExit: A Dynamic Early-Exit Strategy for Deep
  Residual Networks}. In \bibinfo{booktitle}{\emph{IEEE Int. Workshop on Signal
  Processing Systems (SiPS)}}. \bibinfo{pages}{178--183}.
\newblock


\bibitem[Wang et~al\mbox{.}(2019a)]%
        {wang2019see}
\bibfield{author}{\bibinfo{person}{Zizhao Wang}, \bibinfo{person}{Wei Bao},
  \bibinfo{person}{Dong Yuan}, \bibinfo{person}{Liming Ge},
  \bibinfo{person}{Nguyen~H Tran}, {and} \bibinfo{person}{Albert~Y Zomaya}.}
  \bibinfo{year}{2019}\natexlab{a}.
\newblock \showarticletitle{{SEE}: Scheduling early exit for mobile DNN
  inference during service outage}. In \bibinfo{booktitle}{\emph{ACM Int. Conf.
  on Modeling, Analysis and Simulation of Wireless and Mobile Systems}}.
  \bibinfo{pages}{279--288}.
\newblock


\bibitem[Xin et~al\mbox{.}(2020)]%
        {ACL2020_Deebert}
\bibfield{author}{\bibinfo{person}{Ji Xin}, \bibinfo{person}{Raphael Tang},
  \bibinfo{person}{Jaejun Lee}, \bibinfo{person}{Yaoliang Yu}, {and}
  \bibinfo{person}{Jimmy Lin}.} \bibinfo{year}{2020}\natexlab{}.
\newblock \showarticletitle{{D}ee{BERT}: Dynamic Early Exiting for Accelerating
  {BERT} Inference}. In \bibinfo{booktitle}{\emph{Proceedings of the 58th
  Annual Meeting of the Association for Computational Linguistics}}.
  \bibinfo{publisher}{Association for Computational Linguistics},
  \bibinfo{pages}{2246--2251}.
\newblock


\end{thebibliography}


\end{document}